\RequirePackage{fix-cm}
\documentclass[smallextended]{svjour3}       
\usepackage{soul}
\usepackage{xargs}                      
\usepackage{xspace}
\usepackage{xcolor}
\usepackage{svg}
\usepackage{hyperref}
\usepackage{natbib}

\newcommand{\orcid}[1]{\href{https://orcid.org/#1}{\includegraphics[width=10pt]{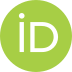}}}
\smartqed  
\usepackage{graphicx}
\begin{document}

\title{Assessing Linguistic Generalisation 
in Language Models
}
\subtitle{A Dataset for Brazilian Portuguese}


\author{Rodrigo Wilkens\textsuperscript{1}\orcid{0000-0003-4366-1215}         \and
        Leonardo Zilio\textsuperscript{2}\orcid{0000-0002-6101-0814}         \and
        Aline Villavicencio\textsuperscript{3}\orcid{0000-0002-3731-9168}
}

\authorrunning{Wilkens et al.} 

\institute{
Rodrigo Wilkens \at 
                    rodrigo.souzawilkens@unimib.it 
\and
Leonardo Zilio \at 
                    l.zilio@surrey.ac.uk 
\and
Aline Villavicencio \at 
                    a.villavicencio@sheffield.ac.uk 
\\ \\ 
1 - University of Milano-Bicocca, Italy \\ \\ 
2 - University of Surrey, United Kingdom \\ \\ 
3 - University of Sheffield, United Kingdom
}

\date{Received: date / Accepted: date} 

\maketitle

\begin{abstract}

Much recent effort has been devoted to creating large-scale language models.  Nowadays, the most prominent approaches are based on deep neural networks, such as BERT. However, they lack transparency and interpretability, and are often seen as blackboxes, which affect their applicability in downstream tasks as well as the comparison of different architectures or even the same model trained on different corpora or hyperparameters.  In this paper, we propose a set of intrinsic evaluation tasks that inspect the linguistic information encoded in models developed for Brazilian Portuguese. These tasks are designed to evaluate how different language models generalise information related to grammatical structures and multiword expressions (MWEs), thus allowing for an assessment of whether the model has learned different linguistic phenomena.
The dataset that was developed for these tasks is composed by a series of sentences with a single masked word and a cue that narrows down the context.
This dataset is divided into MWEs and grammatical structures, and the latter is subdivided into 6 tasks: impersonal verbs, subject agreement, verb agreement, nominal agreement, passive and connectors. 
The subset for MWEs was used to test BERTimbau Large, BERTimbau Base and mBERT. For the grammatical structures, we used only BERTimbau Large, because it yielded the best results in the MWE task. In both cases, we evaluated the results considering the best candidates and the top ten candidates. The evaluation was done both automatically (for MWEs) and manually (for grammatical structures). The results obtained for MWEs show that BERTimbau Large surpassed both the other models in predicting the correct masked element. However, the average accuracy of the best model was only 52\% when only the best candidates were considered for each sentence, going up to 66\% when the top ten candidates were taken into account. 
As for the grammatical tasks, results presented better prediction, but also varied depending on the type of morphosyntactic agreement. Cases such as connectors and impersonal verbs, which do not require any agreement in the produced candidates, had precision of 100\% and 98.78\% among the best candidates, while other tasks that require morphosyntactic agreement to produce good candidates had results consistently below 90\% overall precision, with the lowest scores being reported for nominal agreement and verb agreement, both having below 80\% overall precision among the best candidates. Therefore, we identified that a critical and widely adopted resource for Brazilian Portuguese NLP although mostly proficient in these tests, also presents issues concerning MWE vocabulary and morphosyntactic agreement. These models are the de-facto core component in current NLP systems, and our findings demonstrate the need of additional improvements in these models and the importance of widely evaluating computational representations of language.  

\end{abstract}

\section{Introduction} \label{intro}

Learning computational representations that accurately model language is a critical goal of Natural Language Processing (NLP) research. These representations, or word embeddings, are  not only helpful for language technology tasks like machine translation, question answering and text adaptation. They can also improve our understanding of human cognition, for instance, correlating well with human predictions about words \citep{8502464,Schrimpf2020.06.26.174482}, and also helping to highlight the existence of explicit and implicit social constructs, including gender and racial bias 
\citep{10.5555/3157382.3157584,kumar-etal-2020-nurse}, or
helping to detect misinformation \citep{oshikawa2018survey,su2020motivations} and cultural change \citep{savoldi2021gender,dinan2020multi}, among others. With the recent popularisation of NLP and the availability of large scale computing resources, we have witnessed a quick evolution of many new computational architecture proposals for language representation. As a consequence, there are currently pre-trained language models available in an unprecedented scale, for over a hundred languages \citep{devlin2018bert}, including low-resourced ones. 
However, these models lack transparency and interpretability \citep{DBLP:journals/corr/abs-2002-06177} and are often seen as blackboxes, all of  which limit their application in tasks that require explainability. 
This fast-paced evolution raises the need for careful assessment of the kinds of information that a representation incorporates, both in terms of linguistic and common-sense or world information. Moreover, the evaluation procedures need to adopt standard protocols that are independent of the architecture proposed, and be applicable to classes of models in general, setting the standards for what we expect them to have learned and be proficient in.

In fact, much recent effort has been devoted to defining evaluation protocols that allow us to have an insight into the knowledge that the model incorporates    \citep{ettinger2020bert,warstadt-etal-2020-blimp,vulic-etal-2020-multi}. 
These evaluations may target different phenomena, but their main distinction is between intrinsic and extrinsic evaluations. Intrinsic evaluations usually involve comparing predictions made by a model to human judgements, existing resources like WordNet \citep{miller1990introduction} or psycholinguistic norms.   
Extrinsic evaluations are often based on applications \citep{bakarov2018survey} and examine the ability of  embeddings to be used as the feature vectors of supervised machine learning algorithms for tasks like natural language entailment and question answering. The assumption is that better quality and accurate embeddings lead to better quality applications.  
Recently, the BERT \citep{devlin2018bert} architecture (and its variations, such as RoBERTa \citep{liu2019roberta}, DistilBERT \citep{sanh2019distilbert}, XLNet \citep{yang2019xlnet} and ALBERT \citep{albert}) has been used as a language encoding component in several applications that enhanced the state-of-the-art in their several NLP fields. These developments can be seen as an extrinsic evaluation of the BERT architecture, as they seem to suggest that BERT more successfully encodes syntactic and semantic information of the language than alternative models. However, what these evaluations do not indicate is which specific information is encoded, and if it is used for a given task in ways that are compatible with human use of language. In this paper we focus on intrinsic evaluations with an aim at gaining a better understanding of how linguistically proficient these models really are.    

For different languages, the stage of development and proficiency of the models varies considerably, and so do the datasets available for evaluation. For Portuguese, in particular, only a few evaluation initiatives are available, such as \cite{souza2020bertimbau,smallermbert,schneider2020biobertpt}, 
and, despite being valuable evaluation contributions to NLP in Portuguese, these correspond to only extrinsic evaluations and target very specific tasks. Overall, the Portuguese versions of BERT have not been subjected to much intrinsic evaluations in comparison to their English counterparts, and there is great uncertainty about the coverage and quality of the linguistic information about Portuguese encoded in these models. As a consequence, there is also uncertainty about the stability and quality of the output produced by systems built using these models. In this paper, we address some of these issues, assessing the linguistic generalisation of language models for Portuguese. In particular, we propose a model-agnostic test set to measure the ability of a model to capture expected linguistic patterns found in Brazilian Portuguese. 
The main contributions of this study are:
\begin{itemize}
    \item A dataset for testing generalisation both related to multiword expression (MWE) and grammatical information. The MWE dataset targets 33 non-compositional MWEs, while the set for testing grammatical information is divided into 6 different tasks.
    \item An analysis that generated a language profile of BERT models for Brazilian Portuguese.
    \item A comparison of the BERT models' quality considering only the best generated candidates and the ten best candidates, also highlighting cases where the model puts more confidence on wrong candidates.
\end{itemize}
The proposed dataset will be available on github.

The following sections are organised as follows: Section \ref{sec:related} summarises the most influential model-agnostic proposals for intrinsic evaluation, including those targeting Portuguese. Section \ref{sec:bert} presents the BERT models for Portuguese, describing their characteristics, and the models used in this work. Then we detail the methodology employed to create seven tasks comprised on our test set in Section \ref{sec:methodology}. The results of the different models and their analyses are presented in Section \ref{sec:results}. Finally, in Section \ref{sec:conclusion} we discuss the conclusions of our findings as well as future work.











\section{Related Work} \label{sec:related}

The performance that state-of-the-art models achieve on language tasks is seen as an indication of their ability to capture linguistic information. However, their lack of transparency and interpretability prevents an in-depth analysis of the linguistic patterns and generalisations that they capture. Therefore, considerable attention has been devoted to the development of intrinsic evaluation protocols for inspecting the information captured by these models. 

Intrinsic evaluations usually involve experiments in which word embeddings are compared with human judgements about word relations, and may be divided into various categories. For example, \cite{bakarov2018survey} arranges them as follows:

\begin{itemize}
\item[a] conscious evaluation including tasks like semantic similarity, analogy, thematic fit, concept categorisation, synonym detection, and outlier word detection; 
\item[b] subconscious evaluation in tasks like semantic priming, neural activation patterns, and eye movement data; 
\item[c] thesaurus-based evaluations, including thesaurus vectors, dictionary definition graph, cross-match test, semantic difference and semantic networks; and 
\item[d] language-driven of phonosemantic analysis and bi-gram co-occurrence frequency.
\end{itemize}

In this paper, we concentrate mainly on the conscious evaluation tasks. These vary from testing morphosyntactic agreement patterns, like number and gender \citep{warstadt-etal-2020-blimp}, to more semantic-related information, like the implications of negation \citep{ettinger2020bert,kassner-schutze-2020-negated}. As an approach to evaluate sensitivity to syntactic structures in English, \cite{linzen2016assessing} proposed an evaluation focusing on number agreement for subject-verb dependencies, which is something that we also address in this paper. 
They generated a dataset of 1.35 million number prediction problems based on Wikipedia (9\% for training, 1\% for validation, and 90\% for testing). These problems were used to evaluate four tasks: 
\begin{itemize}
\item[a] number prediction, with a binary prediction task for the number of a verb; 
\item[b] verb inflection, with a variation of the number prediction when the singular form of the upcoming verb is also given to the model; 
\item[c] grammaticality judgements as another classification task, but the model must indicate the grammaticality of a given sentence; and 
\item[d] language modelling in which the model must predict a correct word with the highest probability than a wrong word in a given sentence.
\end{itemize}

A related dataset, the ``colorless green ideas'' test set, was proposed by \cite{gulordava2018colorless} including four languages (Italian, English, Hebrew, Russian). The test items focus on long-distance number agreement evaluation, evaluating the accuracy in cases of subject-verb agreement with an intervening embedded clause and agreement between conjoined verbs separated by a complement of the first verb. Their test set is composed of syntactically correct sentences from a dependency treebank which are converted into nonce sentences by replacing all content words with random words with the same morphology, aiming to avoid semantic cues during the evaluation. 

To avoid very implausible sentences that violate selectional restrictions (e.g., the apple laughs), \cite{marvin2018targeted} present an automatically constructed dataset for evaluating the grammaticality of the predictions of a language model. \cite{marvin2018targeted} use templates to automatically generate 350,000 English sentence pairs, consisting of pairs of grammatical and ungrammatical sentences, for examining subject-verb agreement, reflexive anaphora and negative polarity.

To evaluate the performance obtained by BERT in 
these datasets \citep{gulordava2018colorless,linzen2016assessing,marvin2018targeted}, \cite{goldberg2019assessing} fed a masked version of a complete sentence into BERT, then compared the score assigned to the original correct verb to the score assigned to the incorrect one. 

\cite{mueller2020cross} evaluated LSTM language models and the monolingual and multilingual BERT versions on the subject-verb agreement challenge sets. They used CLAMS (Cross-Linguistic Assessment of Models on Syntax), which extends \cite{marvin2018targeted} to include English, French, German, Hebrew and Russian. 
To construct their challenge sets, \cite{mueller2020cross} use a lightweight grammar engineering framework (attribute-varying grammars), aiming at more flexibility than the hard-coded templates of \cite{marvin2018targeted}. One may argue that this approach is related to ours because they use a grammar with attributes to guide the sentence generation while our work uses pre-selected seeds. Our seeds are related to their attributes, and our generation of grammatical sentences is the combination of seed and original target word while the ungrammatical sentences may be generated by a cross-combination of seed and original target word. 
Their experiments on English BERT and mBERT suggest that mBERT seems to learn syntactic generalisations in multiple languages, but not in the same way in all languages. In addition, its sensitivity to syntax is lower than that of monolingual BERT.

Regarding evaluations for Portuguese, \cite{bacon2019does} evaluated BERT's sensitivity to four types of structure-dependent agreement relations for 26 languages, including Portuguese. They showed that both the monolingual and multilingual BERT models capture syntax-sensitive agreement patterns. \cite{bacon2019does} evaluated the models in a cloze task, in which target words would share the morphosyntactic features of the word with which they agree. 
In the cloze task, the data comes from version 2.4 of the Universal Dependencies (UD) treebanks \citep{nivre2017universal}, using the part-of-speech and dependency information to identify potential agreement relations. 
BERT responses are then annotated with morphosyntactic information from both the UD and the UniMorph projects \citep{sylak2016composition} to be compared with the morphosyntactic features of the word with which they agree. With respect to Portuguese, \cite{bacon2019does} evaluate 47,038 sentences in the cloze test and 2,107 in the feature bundles using mBERT, and the results obtained show that it performed remarkably well, achieving an accuracy close to 100\%. 

In another multilingual evaluation, \cite{csahin2020linspector} introduced 15 probing tasks at type level for 24 languages, finding that some probing tests correlate to downstream tasks, especially for morphologically rich languages. 
They performed an extrinsic evaluation using downstream tasks, assessing the performance in POS tagging, dependency parsing, semantic role labelling, named entity recognition, and natural language inference targeting German, Russian, Turkish, Spanish and Finnish. 
Moreover, they propose the contextless prediction of the morphological feature of a given word, aiming to identify the feature indicated in the UniMorph dictionary \citep{sylak2016composition}. For that, \cite{csahin2020linspector} removed ambiguous forms and partially filtered infrequent words. This evaluation included features like the following for Portuguese: number, gender, person, 
and tense. 
However, they did not report the models' performance for Portuguese. 

\section{BERT models for Portuguese} \label{sec:bert}

Large pre-trained language models are valuable assets for language processing. These models support technological and scientific improvements. For English, for example, we can easily find 2,795 models accessing Huggingface website\footnote{\url{https://huggingface.co/models?filter=en}}. However, this abundance of models is not true for most languages. For example, there are only 67 models for Portuguese available in the same repository\footnote{\url{https://huggingface.co/models?filter=pt}}, and most of these are for Machine Translation and Automatic Speech Recognition, and Text2Text Generation. In this section, we highlight three models dedicated to language processing: smallermBert and BERTimbau, which are based on Portuguese for general purposes, and BioBERTpt, which is based on Portuguese for special Purposes.

Improving the mBERT model, \cite{smallermbert} proposed to generate smaller models based on mBERT that are language-specific. In other words, remove the parameters related to the non-target language. Then, they identified the vocabulary of each language and rebuilt the embedding layer to generate the corresponding models. 
Following the original mBERT, \cite{smallermbert} started from the entire Wikipedia dump of each language, initially covering 15 languages, but since the work's publication, they have been extending it to other languages, including Portuguese\footnote{\url{https://huggingface.co/Geotrend/bert-base-pt-cased}}.

In a different perspective, \cite{souza2019portuguese,souza2020bertimbau} trained language-specific models for Brazilian Portuguese (nickname BERTimbau), using data from brWaC~\citep{wagner2018brwac}. 
Their models were trained over 1,000,000 steps using the BERT-Base and BERT-Large Cased variants architectures. In order to evaluate their models, they resorted to three probing tasks (sentence textual similarity, recognising textual entailment, and named entity recognition) in which BERTimbau improved the state-of-the-art compared to multilingual models and previous monolingual approaches.

Targeting clinical NLP, \cite{schneider2020biobertpt} trained BioBERTpt, a deep contextual embedding model for Portuguese, by fine-tuning mBERT. BioBERTpt is trained using clinical narratives and biomedical-scientific papers in Brazilian Portuguese.
\cite{schneider2020biobertpt} trained three versions of BioBERTpt: BioBERTpt(bio) trained only on biomedical literature from scientific papers from Pubmed and Scielo; BioBERTpt(clin) trained on clinical narratives from electronic health records from Brazilian Hospitals; and BioBERTpt(all) trained on clinical narratives and biomedical literature in the Portuguese language.

Aiming to evaluate the BERT models for Brazilian Portuguese, we selected two models that were trained using brWaC~\citep{wagner2018brwac} and multilingual BERT, allowing a comparison point with BERT models trained in other languages. We used both the BERTimbau Base and Large models~\citep{souza2019portuguese,souza2020bertimbau}, which have, respectively, 12 layers and 110M parameters, and 24 layers and 335M parameters. 
These models were chosen because they are uniquely trained in Brazilian Portuguese, thus avoiding issues from different variations, and they are also widely adopted, being the most popular models for Portuguese on the Huggingface webpage (BERTimbau Base was downloaded 17k times and BERTimbau Large was downloaded 11k times). Therefore, the findings of this paper may have a wide impact in downstream applications for Brazilian Portuguese.





\section{Methodology} \label{sec:methodology}
\begin{figure}
\centering
  \includegraphics[width=\textwidth]{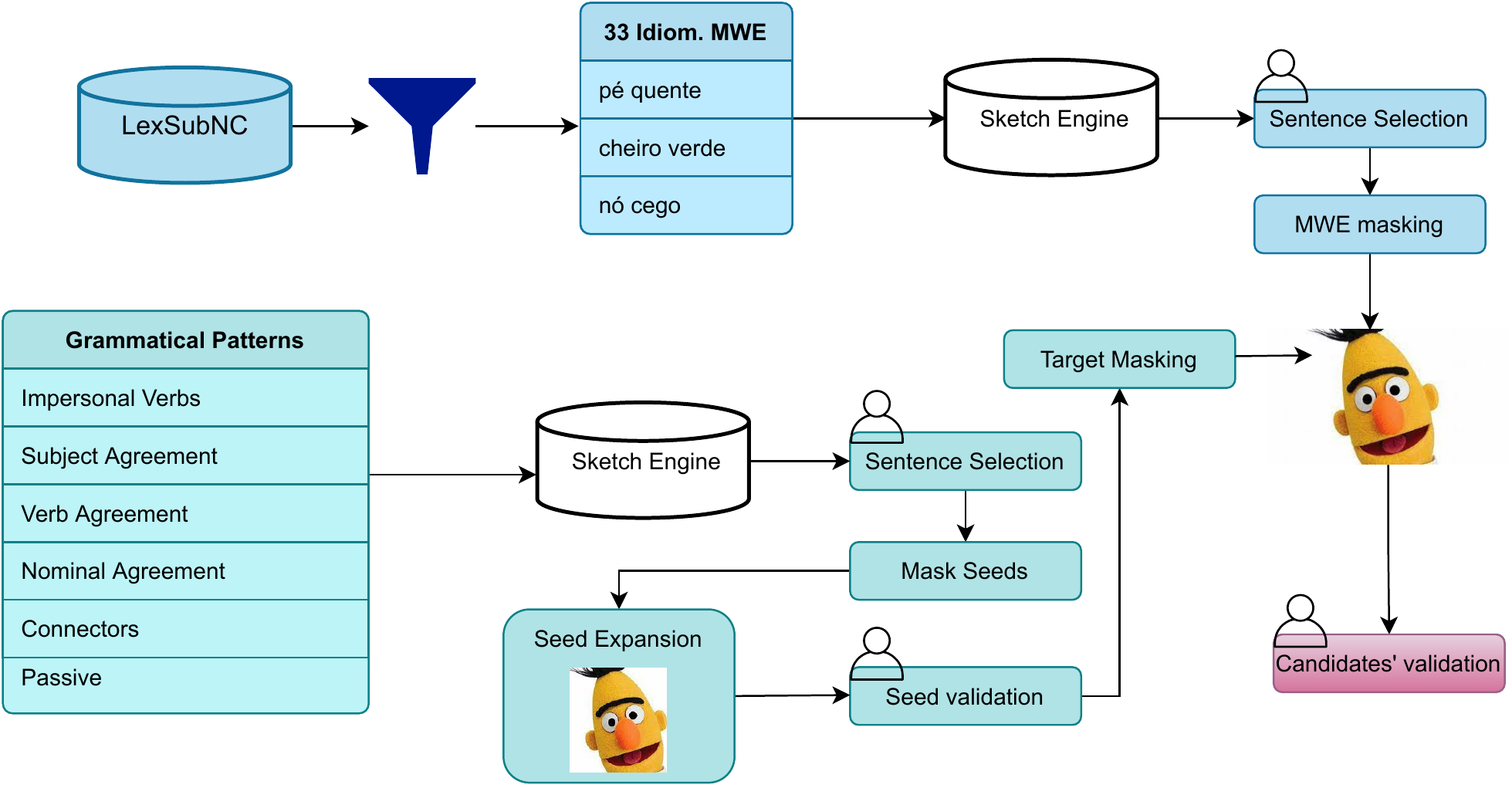}
\caption{Methodology for creating the test dataset.}
\label{fig:metodologia}       
\end{figure}

In this section, we describe the steps performed for the development of the dataset proposed in this work. We first compiled the MWE test set (see Subsection~\ref{sec:results:methodology:mwe}) by selecting idiomatic MWEs and five context sentences for each of them. For the grammar tests (Subsection~\ref{sec:results:methodology:syn}) we also selected context sentences, but we looked for specific grammar patterns in this step. Once we obtained the sentences for both tasks, we created seeds for the sentences, aiming to use them as cues to the specific target. Finally, we fed BERT model with our evaluation sentences, and manually evaluate the model's output (see Section~\ref{sec:results}). An overview of this methodology is presented on Figure \ref{fig:metodologia}. 

\subsection{Multiword Expression Tests} \label{sec:results:methodology:mwe}
Multiword Expressions (MWEs), like nominal compounds (NCs) and idioms, can be defined as linguistic structures that cross word boundaries \citep{sag2002multiword}. Due to their many potential linguistic and statistical idiosyncrasies \citep{sag2002multiword}, knowledge about MWEs is often considered a mark of proficiency for language learners. 

In the MWE task, our goal is to assess the MWE inventory of a model by  measuring the models' ability to identify an MWE given a context. For that, we use 33 highly idiomatic two-word NCs\footnote{The NCs were annotated by human judges using a Likert scale from 0 (idiomatic) to 5 (compositional). We chose NCs whose average compositionality scores were lower than 1.} from LexSubNC \citep{wilkens:2017}, such as \textbf{pão duro} (stingy person; lit. \emph{hard bread}) and \textbf{pé quente} (person who brings good luck; lit. \emph{hot foot}), and then we select 5 sentences from Corpus Brasileiro \citep{sardinha2010corpus} for each MWE. 
The sentence selection was done by shuffling concordance results on Sketch Engine \citep{kilgarriff04sketch} and selecting 5 full sentences of different lengths. This process resulted in 165 sentences for testing MWEs.

For assessing 
the models' capacity to retrieve the target NC given a compatible context, we adopt a masking protocol, in which we feed the model an input sentence using a mask to replace one word of the NC. The model output is then evaluated in terms of whether or not the model produces the correct word for replacing the mask. Although the mask expect only one possible answer, different words may also be used to replace the mask, and we analyse if the target word is among the responses produced by the model. Moreover, as the NCs are composed of two words (and adjective and a noun), for each NC we generated two test items, keeping one of the NC components as cue for the model and masking the other one. 
For example, the sentence \textit{Presidente, trago uma notícia em \textbf{primeira mão}.} (\emph{President, I bring you first-hand news.}), which contains the NC \textit{primeira mão} (\emph{first-hand}), is processed twice: 
\begin{itemize}
    \item \textit{Presidente, trago uma notícia em [MASK] \textbf{mão}.}
    \item \textit{Presidente, trago uma notícia em \textbf{primeira} [MASK].}
\end{itemize}
We expect that together, these sentences, along with one of the NC components, should provide enough context about the NC to trigger the model to produce the missing word. The output of this process resulted in 10 candidate words for each test item, in other words, 20 candidate words for each MWE, with 10 for the first component word of the compound and 10 for the second.

\subsection{Grammatical Tests} \label{sec:results:methodology:syn}

To assess the models' 
grammatical information, we developed a broad range of cloze tests, each one targeting a specific phenomenon, with several sub-phenomena, where each sub-phenomenon was tested using a set of sentences and a series of variations. The targets of these tests were: impersonal verbs, subject agreement, verb agreement, nominal agreement, passive and connectors. 
These test were inspired by the work of \cite{kurita2019quantifying}, who propose a template-based method to quantify social bias in BERT.
Each test sentence is composed of a mask and a set of seeds as parameters. Like in the MWE test, the mask is the ``blank", and the seeds work as a linguistic cue to the expected answer. For example, in the nominal agreement test, the sentence \textit{``\'{E} precisamente na revela\'{c}\~{a}o \textit{\textless SEED\textgreater [MASK]} que reside o valor dessas cartas, diz.''} (\emph{It is precisely in the revelation \textit{\textless SEED\textgreater [MASK]} that lies the value of those letters, he/she says.}) is used 40 times, each one using a different seed such as 
\textit{das rela\'{c}\~{o}es} (\emph{of the relations}), \textit{das capacidades} (\emph{of the capacities}),
\textit{da alma} (\emph{of the soul}), \textit{da condi\'{c}\~{a}o} (\emph{of the condition}), 
\textit{dos seres} (\emph{of the beings}), \textit{dos segredos} (\emph{of the secrets}), 
\textit{do ser} (\emph{of the being}), and \textit{do esp\'{i}rito} (\emph{of the spirit}). 
In this example, we use four types of seeds each one targeting a different output in the masked work (male singular, male plural, female singular and female plural). 
The different seed sets are specific for each sentence, 
because they have to take into consideration its syntactic and semantic restrictions. 

For nominal agreement, we automatically generated seed sets using a mask where the original seed would be, this allowed us to have more extensive coverage. In this process, we selected the top 10 candidates from base-BERT and then we manually evaluated the candidates to eliminate bad fits (i.e., we removed the candidates with agreement errors or a different meaning). For verb and subject agreement, and for impersonal verbs, we automatically generated a set of seeds based on the UNITEX-PB dictionary \citep{muniz2005unitex,vale2015novo}, and then also validated each seed in context. In the case of passive voice, due to the complexity of the seeds, we had to manually generate sets of seeds for each sentence. Finally, in the case of connectors, we could not use any type of seed, because this would require us to generate full clauses, so we only used five sentences per connector that was tested.

Using the manually validated templates composed of sentences and their seeds, we generated 10 candidate answers for each masked word using BERTimbau Large\footnote{Given the size of the manual evaluation task, with more than 12k items, we only used the BERTimbau Large model, as it performed best in the MWE task, as will be discussed in the next section.}. These candidates were also annotated with part-of-speech using the Unitex-PB dictionary \citep{muniz2005unitex,vale2015novo}.\footnote{We do not use a parser in this step for simplicity.} Then, all candidate words from BERTimbau Large were evaluated by a linguist according to their syntactic and semantic suitability in context. 
Although we used seeds as cues for the expected type of answer (for instance, we used generic passive structures to induce the model to produce a nominal participle as candidate in the passive voice tests), the actual answer of the system could be from a different category, and still the sentence would be grammatically and semantically correct. As such, the evaluation took into account any answer that would fit in the context, not necessarily only the ones that were expected. The ones that were correct, but deviated from the expected target answer, were identified, and are further discussed in the results.

In total, the dataset for grammatical evaluation consists of 6 dimensions that contain 
1,231 tests and 688 seeds. 

\section{Results} \label{sec:results}

This section presents the evaluation results considering the two different subsets of tests: MWEs and grammatical information. The MWE results consider three models: BERTimbau Large, BERTimbau Base and mBERT. These models were evaluated in terms of accuracy of the best prediction and accuracy among the top ten candidates. The grammatical evaluation was done on BERTimbau Large, and considered results in terms of precision of the best candidate (or accuracy) and precision at the top ten candidates.

\subsection{MWE Tests} \label{sec:results:mwe}

Aiming to assess the model generalisation concerning MWE vocabulary, we calculate the accuracy of the output. However, this information might not be fully representative of its generalisation capacities. Thus, for each masked sentence we looked for the correct word in the top ten predictions of the model. In the evaluation step, we resorted to accuracy at ten (acc@10). In other words, we evaluated the presence or absence of the expected word in the top ten predictions.

The multilingual model, mBERT, performed consistently worse than the dedicated model trained for Portuguese. 
Moreover, BERTimbau Large performs better than the base version. In both these cases, it was the larger models that performed better, suggesting that the ability to learn MWEs is related to the size of the model. 
In terms of the difficulty of the task, as shown in Table \ref{tab:acc_mwe}, apart from the multilingual BERT, the models were able to predict the missing MWE component as the first alternative in 40-52.73\% of the cases. However, using this more strict evaluation scenario where only the top choice is considered, the quality of the prediction is poor. In a more lenient scenario, When we analyse the ability of a model to predict the missing word among the top 10 most likely words, the results substantially improve. While mBERT has an average increase of 9\% 
accuracy in comparison with only the top prediction, the BERTimbau models have a much more substantial increase: BERTimbau Base has an improvement of 17.58\% in relation to the top candidate and BERTimbau Large of 15.75\% from 51.52\% to 67.27\% accuracy. Although mBERT shows good capacities in different NLP tasks, the model captured little to no information to predict idiomatic items that are specific to Brazilian Portuguese. Additionally, we observed that the gain of including more candidate words is mainly restricted to the top two candidates. 

The changes from BERTimbau Base to BERTimbau Large allowed the model to learn a larger MWE inventory, considering the target MWEs, and have more accurate prediction given the clues in the masking task (Table \ref{tab:mwe_error}). However, there were two cases in which performance decreased using the larger model: \emph{sangue azul} (blue blood) and \emph{p\~{a}o duro} (stingy person; lit. \emph{hard bread}); and other cases that are still not accurately represented by either model (Table \ref{tab:mwe_error2}). Overall, the difference in performance between BERTimbau Large and Base was not as big as the difference between them and mBERT, but BERTimbau Large was able to learn more MWE and displayed more confidence in correct predictions. However, we also noticed that the values of accuracy at ten for BERTimbau Base were similar to BERTimbau Large, indicating that exploring more outputs from the Base model might have the same performance of BERTimbau Large, which require more processing power.


\begin{table}[]
\caption{Accuracy of different sentences for Word1 and Word2. acc@10 means the accuracy considering the 10 most likely candidates. For example, if we consider only the word with the highest probability from BERTMbau-Base, it achieves an accuracy of 40\%, however, if we consider all 10 candidates, evaluating only if the correct one is on the list, it goes to 57.58\%}
\centering
\begin{tabular}{|l|c|r|r|}
\hline
\multicolumn{1}{|c|}{\textbf{Model}} & \textbf{Word} & \multicolumn{1}{c|}{\textbf{ACC}} & \multicolumn{1}{c|}{\textbf{ACC@10}} \\ \hline
BERTimbau Base                       & 1             & 40.00\%                           & 57.58\%                              \\ \hline
BERTimbau Base                       & 2             & 45.45\%                           & 64.24\%                              \\ \hline
BERTimbau Large                      & 1             & 52.73\%                           & 65.45\%                              \\ \hline
BERTimbau Large                      & 2             & 51.52\%                           & 67.27\%                              \\ \hline
mBERT                                & 1             & 6.67\%                            & 16.97\%                              \\ \hline
mBERT                                & 2             & 3.64\%                            & 11.52\%                              \\ \hline
\end{tabular}
\protect\label{tab:acc_mwe}
\end{table}

\begin{table}[] 
\centering
\caption{Performance comparison between BERTimbau Large and BERTimbau Base. MWEs which were predicted by the BERTimbau Large, but not by BERTimbau Base.}
\begin{tabular}{|l|l|}\hline
{\textbf{Word 1}} & {\textbf{Word 2}} \\ \hline
 livro aberto               	&	 livro aberto               	\\\hline
 montanha russa             	&	 montanha russa             	\\\hline
 n\'o cego                    	&	 olho mágico                	\\\hline
 olho m\'agico                	&	 pau mandado                	\\\hline
 p\~ao duro                   	&	 pavio curto                	\\\hline
 p\'e frio                    	&	 p\'e frio                    	\\\hline
 p\'e quente                  	&	 p\'e quente                  	\\\hline
 peso morto                 	&	 peso morto                 	\\\hline
 planta baixa               	&	 planta baixa               	\\\hline
 sangue azul                	&	 saia justa                 	\\\hline
 sangue frio                	&	 sangue frio    	\\   \hline         
\end{tabular}
\protect\label{tab:mwe_error}
\end{table}

\begin{table}[] 
\centering
\caption{Performance comparison between BERTimbau Large and BERTimbau Base: MWEs were not predicted by either model.}
\begin{tabular}{|l|l|}\hline
{\textbf{Word 1}} &{\textbf{Word 2}} \\ \hline
bode expiat\'orio            	&	 bode expiat\'orio            	\\\hline
cheiro verde               	&	 gato pingado               	\\\hline
elefante branco            	&	 longa metragem             	\\\hline
ovelha negra               	&	 nó cego                    	\\\hline
pavio curto                	&	 olho gordo                 	\\\hline
pente fino                 	&	 pente fino                 	\\\hline
roleta russa               	&	 vista grossa               	\\\hline
saia justa                 	&	         \\\hline                   	              
\end{tabular}
\protect\label{tab:mwe_error2}
\end{table}

\subsection{Grammatical Tests} \label{sec:results:syn}

This section presents the results of the grammatical tests to which BERTimbau Large was submitted. As there were six very different grammatical tests, we report and discuss their results individually as precision at 1 (or accuracy) and precision at 10, aiming to analyse the model's proficiency. At the end of this section, we make a brief comment on the overall performance of the model.

\subsubsection{Impersonal Verbs}

The set of sentences with impersonal verbs tested whether the model would produce candidates that were not subjects. The verbs used as cue represented meteorological verbs (such as to rain, to snow) and existential verbs (such as to exist, to have been), which do not accept any subject, and are therefore defective in their conjugation. As such, we expected the model to produce answers that were not nouns or pronouns.

As we can see in Table \ref{tab:impverbs}, the models perform well in this task, with  meteorological verbs having slightly worse results, as the model still produced some answers that did not fit. Results considering the top candidate were near 100\%, and the average precision 
considering the top 10 candidates decreases to 76.04\%. For  existential verbs, results were much higher, with a precision of 97.50\% among the top 10 candidates and 100\% for the top 1. The model was able to generate punctuation marks, such as parenthesis or quotation marks, in most of the templates, which were a good fit for the test sentences.

Looking at the different tenses, to check whether there is any impact of the verb form used as a cue for the model, we see that only the pluperfect form produced results below 100\% for the top candidate, and was also the worse cue on the precision at 10 evaluation, with 65.33\%. This could be a reflect of the fact that the pluperfect tense in Brazilian Portuguese is usually written in a compound form, using ``ter'' (have) as auxiliary verb, conjugated in the imperfect tense, associated with the past participle of the main verb, so that the pluperfect simple form, which was the form used in the test sentences, is not widely used anymore. Assuming that lower frequencies have an impact on the quality of the representation, they might not have been learned well by the model. Nonetheless, the performance confirms the proficiency of the model with impersonal verbs.

\begin{table}[]
\caption{Impersonal verbs: Verb types}
\centering
\label{tab:impverbs}
\begin{tabular}{|l|l|l|}
\hline
\multicolumn{1}{|c|}{\textbf{Case name}} & \multicolumn{1}{c|}{\textbf{P@1}} & \multicolumn{1}{c|}{\textbf{P@10}} \\ \hline
Meteorological Verbs                                   & 97.56\%                           & 76.04\%                            \\ \hline
Existential Verbs                                   & 100.00\%                          & 97.50\%                            \\ \hline
\end{tabular}
\end{table}

\begin{table}[]
\caption{Impersonal verbs: Results by tense and mode}
\centering
\label{tab:impverbs_tenses}
\begin{tabular}{|l|l|l|}
\hline
\multicolumn{1}{|c|}{\textbf{Case name}}                  & \multicolumn{1}{c|}{\textbf{P@1}} & \multicolumn{1}{c|}{\textbf{P@10}} \\ \hline
Past Indicative          & 100.00\%                              & 82.00\%                             \\ \hline
Present Indicative                           & 100.00\%                              & 85.63\%                             \\ \hline
Future Indicative         & 100.00\%                              & 80.00\%                             \\ \hline
Imperfect Indicative                & 100.00\%                              & 95.71\%                             \\ \hline
Pluperfect Indicative & 86.67\%                               & 65.33\%                             \\ \hline
Future Subjunctive       & 100.00\%                              & 79.00\%                             \\ \hline
\end{tabular}
\end{table}

\subsubsection{Subject Agreement}

For the subject agreement task, the model was expected to produce a subject that would agree in person and number with the provided verb seeds. Given that Portuguese allows for hidden or indeterminate subject, we expected that the system would produce some results that would not fit as subjects, but that would fit in the sentences.

Considering the type of subject that was expected given the verb conjugation, as seen in Table \ref{tab:subjagreepers}, results both in the top 1 and in the top 10 were varied, ranging from 64.10\% for the second person singular to 100\% for the third person singular. The model had a hard time to produce good candidates when the expected subject should fit a second person singular or plural, which are less commonly used conjugations, but it is interesting to see that the model has higher confidence in wrong answers when we look at the results for the third person plural.

When we look at the tense and mode of the conjugated seed (Table \ref{tab:subjagreetense}, results also varied, and it is interesting to notice that the tense that yielded worst results in the top 1 candidates (barely above 75\% precision) was present indicative, which is one of the most common tenses, while future indicative was the tense with best results (above 92\% precision). Among the top 10 results, pluperfect indicative was the worst, with 71.78\%, and the best result was again the future indicative, with above 90\% precision.

To sum up, the model has a fairly precise capacity of generating subjects that fit in the sentences, but some verb conjugations (second persons, and third person plural) and tenses (present and pluperfect indicative) proved to be a challenge.

\begin{table}[]
\caption{Subject agreement: Results by expected person and number}
\centering
\begin{tabular}{|l|l|l|}
\hline
\multicolumn{1}{|c|}{\textbf{Case name}} & \multicolumn{1}{c|}{\textbf{P@1}} & \multicolumn{1}{c|}{\textbf{P@10}} \\ \hline
First Person Singular                                 & 97.06\%                           & 93.46\%                            \\ \hline
Second Person Singular                                 & 64.81\%                           & 61.73\%                            \\ \hline
Third Person Singular                                & 100.00\%                          & 95.85\%                            \\ \hline
First Person Plural                                & 94.59\%                           & 92.33\%                            \\ \hline
Second Person Plural                                & 66.67\%                           & 63.70\%                            \\ \hline
Third Person Plural                                    & 85.00\%                           & 93.52\%                            \\ \hline
\end{tabular}
\protect\label{tab:subjagreepers}
\end{table}

\begin{table}[]
\caption{Subject agreement: Results by tense and mode}
\centering
\begin{tabular}{|l|l|l|}
\hline
\multicolumn{1}{|c|}{\textbf{Case name}}       & \multicolumn{1}{c|}{\textbf{P@1}} & \multicolumn{1}{c|}{\textbf{P@10}} \\ \hline
Present Indicative                            & 75.61\%                           & 75.13\%                            \\ \hline
Past Indicative          & 78.72\%                           & 87.10\%                            \\ \hline
Pluperfect Indicative & 81.25\%                           & 71.78\%                            \\ \hline
Future Indicative          & 92.86\%                           & 90.46\%                            \\ \hline
Past Tense Future Indicative        & 82.05\%                           & 79.78\%                            \\ \hline
Imperfect Indicative                & 79.49\%                           & 75.28\%                            \\ \hline
\end{tabular}
\protect\label{tab:subjagreetense}
\end{table}

\subsubsection{Nominal Agreement}

For the task of nominal agreement, we looked into the adjective agreement using a noun as cue. Since Portuguese adjectives agree with nouns in gender and number, we had four different test categories: masculine singular, masculine plural, feminine singular and feminine plural. 

The results (Table \ref{tab:nomagreem} shows that results were good across all categories, reaching up to 93.22\% in masculine plural when all 10 candidates were considered for each mask. Here we see that the suggestion in which the model has most confidence is not always the best, as the results on p@1 were consistently worse then in the top 10, especially for feminine singular, which achieved only 78\% precision on the top 1 candidates. Nonetheless, the model displays fairly good proficiency in nominal agreement.


\begin{table}[]
\caption{Nominal agreement: Gender and number}
\label{tab:nomagreem}
\centering
\begin{tabular}{|l|l|l|}
\hline
\multicolumn{1}{|c|}{\textbf{Case name}} & \multicolumn{1}{c|}{\textbf{P@}} & \multicolumn{1}{c|}{\textbf{P@10}} \\ \hline
Feminine Singular                                    & 78.00\%                               & 92.00\%                             \\ \hline
Masculine Singular                                    & 86.84\%                               & 91.58\%                             \\ \hline
Feminine Plural                                    & 87.18\%                               & 88.46\%                             \\ \hline
Masculine Plural                                    & 91.89\%                               & 93.22\%                             \\ \hline
\end{tabular}
\end{table}

\subsubsection{Verb Agreement}

The task of verb agreement was designed to check whether the model can produce verb candidates that correctly agree with the cue. The sentences used for this test had temporal cues or specific language patterns that would induce or require certain verb conjugations. As seeds, we used pronouns and nouns in singular and plural, but also used some verb structures to check for the production of infinitives and gerunds.

Table \ref{tab:verbagreem} shows the results for each expected verb form. Indicative forms (first two rows) had much better results than subjunctive forms (last three rows), while the non-conjugated forms (rows 3 and 4) had the best results overall, reaching up to 100\% precision considering the top 1 candidate. 

In this specific case, it was observed that some cues were not as effective as others, so we investigated the results based on the different pronouns and nouns that were used as cues. In Table \ref{tab:verbagreem_seeds}, pronouns such as ``tu'' (\emph{you singular}), ``n\'{o}s'' (\emph{we}) and ``v\'{o}s'' (\emph{you plural}) presented a much worse cue, as the model did not seem to be able to produce forms that agree with them. Even when considering only the top 1 candidates, ``tu'' had a result barely above 50\%, while ``v\'{o}s'', a very formal  and infrequent pronoun, did not reach 10\% of precision. While ``tu'' and ``v\'{o}s'' are pronouns that are less commonly used in Brazilian Portuguese, being frequently substituted, respectively, by ``voc\^{e}'' and ``voc\^{e}s'', it is hard to explain why the model does not produce good candidates for a cue like ``n\'{o}s''. One possibility is that this form may 
be replaced by a more informal option (\emph{a gente}, we; lit. \emph{the people}), which uses a conjugation that is homograph with the third person singular.  

Although responses with the expected verb tenses were induced in most cases, 5.00\% of the responses provided by the model were correct, meaning that they fitted well in the sentence, but were not verbs or did not agree with the cue provided.

\begin{table}[]
\caption{Verb agreement: Different expected tenses and modes}
\label{tab:verbagreem}
\centering
\begin{tabular}{|l|l|l|}
\hline
\multicolumn{1}{|c|}{\textbf{Case name}} & \multicolumn{1}{c|}{\textbf{P@1}} & \multicolumn{1}{c|}{\textbf{P@10}} \\ \hline
Past Indicative                                   & 87.51\%                           & 78.44\%                            \\ \hline
Present or Future Indicative                                   & 94.63\%                           & 68.59\%                            \\ \hline
Infinitive                                   & 100.00\%                          & 96.10\%                            \\ \hline
Gerunds                                   & 94.81\%                           & 74.88\%                            \\ \hline
Present Subjunctive                                   & 58.00\%                           & 33.40\%                            \\ \hline
Past Subjunctive                                   & 62.78\%                           & 25.08\%                            \\ \hline
Future Subjunctive/Conditional                                   & 68.15\%                           & 57.10\%                            \\ \hline
\end{tabular}
\end{table}

\begin{table}[]
\caption{Verb Agreement: Breakdown of pronouns}
\label{tab:verbagreem_seeds}
\centering
\begin{tabular}{|l|l|l|}
\hline
\multicolumn{1}{|c|}{\textbf{Case name}} & \multicolumn{1}{c|}{\textbf{P@1}} & \multicolumn{1}{c|}{\textbf{P@10}} \\ \hline
Eu (\emph{I})                                 & 80.00\%                           & 52.67\%                            \\ \hline
Tu (\emph{You singular})                                & 51.61\%                           & 34.84\%                            \\ \hline
Ele/Ela (\emph{He/She})                                & 79.41\%                           & 60.59\%                            \\ \hline
N\'{o}s (\emph{We})                                & 30.77\%                           & 13.08\%                            \\ \hline
V\'{o}s (\emph{You plural})                               & 7.69\%                            & 6.92\%                             \\ \hline
Eles/Elas (\emph{They})                               & 85.71\%                           & 44.29\%                            \\ \hline
\end{tabular}
\end{table}

\subsubsection{Connectors}

In this task the goal was to check whether the model could produce cohesive elements to link sentences together. We used specific connectors, which are shown in Table \ref{tab:connectors},  as seeds for selecting the original set of five sentences for each connector. In this task we had no cues for an expected connector, as usually it is the connectors that establish a meaningful relation between clauses, either by coordination or subordination. This means that, although we had an original connector in the test sentences, the evaluation accepted as correct other forms of cohesion that could change the semantics of the sentence, as long as they produced a sentence with meaning.

As Table \ref{tab:connectors} shows, the model was able to predict connectors with very good precision, reaching 100\% in all cases among the top 1 candidates, and then varying in precision among the top 10 candidates. In terms of correct candidates, 10.77\% were not conjunctions in the traditional sense, but most of these were textual connectors, such as ``finalmente'' (\emph{finally}), ``também'' (\emph{also}). Some of the sentences had a very specific requirement in terms of connector, such as the ones with ``ora'' (now), which is a dual connector (``ora..., ora'' $\sim$ \emph{now..., now}), and thus had no margin for many other connector options, which explains the poor precision among the top 10 candidates in some cases.

The results obtained suggest that these models can proficiently use connectors.   

\begin{table}[]
\caption{Connector Prediction}
\label{tab:connectors}
\centering
\begin{tabular}{|l|l|l|}
\hline
\multicolumn{1}{|c|}{\textbf{Case name}} & \multicolumn{1}{c|}{\textbf{P@1}} & \multicolumn{1}{c|}{\textbf{P@10}} \\ \hline
Caso (\emph{If})                               & 100.00\%                          & 26.67\%                            \\ \hline
Conforme (\emph{According to})                           & 100.00\%                          & 46.67\%                            \\ \hline
Contudo (\emph{However})                            & 100.00\%                          & 86.67\%                            \\ \hline
Enquanto (\emph{While})                            & 100.00\%                          & 66.67\%                            \\ \hline
Nem (\emph{Nor})                                & 100.00\%                          & 50.00\%                            \\ \hline
Ora (\emph{Now})                                & 100.00\%                          & 20.00\%                            \\ \hline
Pois (\emph{Because})                                & 100.00\%                          & 50.00\%                            \\ \hline
Porque (\emph{Because})                             & 100.00\%                          & 40.00\%                            \\ \hline
Portanto (\emph{Therefore})                            & 100.00\%                          & 90.00\%                            \\ \hline
Quando (\emph{When})                             & 100.00\%                          & 26.67\%                            \\ \hline
Se (\emph{If})                                 & 100.00\%                          & 50.00\%                            \\ \hline
Todavia (\emph{However})                            & 100.00\%                          & 96.67\%                            \\ \hline
\end{tabular}
\end{table}

\subsubsection{Passive}

Passive voice in Portuguese has an important characteristic, shared with other Romance languages, which is a somewhat long distance agreement of the nominal participle with the subject. This agreement is illustrated in the following example: \emph{A escolha foi feita.} (\emph{The choice was made.}; lit.\emph{The$_{FemSing}$ choice$_{FemSing}$ was made$_{FemSing}$}). This is different to what happens, for instance, with compound verb tenses, such as the compound pluperfect, where the verb participle of the main verb is used and thus there is no requirement for agreement. An example of this can be seen in this example: \emph{Ela tinha escolhido aquela cor.} (\emph{She had chosen that colour.}; lit. She$_{FemSing}$ had chosen$_{NoAgreement}$ that$_{FemSing}$ colour$_{FemSing}$).

To test this case, we used 13 sentences with a varied number of cues that were made up by verbal constructions with the verb ``ser'' (\emph{to be}). Results in Table \ref{tab:passive} show that the model was able to produce correct candidates most of the time, with 86.65\% precision among the top 1 candidates and 78.38\% among the top 10 candidates.

Interestingly, among the candidates that were not correct, 44.99\% were a participle, but had incorrect nominal agreement with the subject of the sentence. The results in the table also point to the model having trouble producing good candidates for the feminine cases, in particular for the singular form, since  for the plural it had better confidence in the correct candidates.

Finally, within the list of correctly generated candidates, not many deviated from the expected word form, as only 5.07\% of the candidates were adjectives that fit the context resulting in a grammatically correct option, instead of the target nominal participle.

For this test too, the models are  proficient in generating the agreement for the passive form for most cases apart from the feminine singular.


\begin{table}[]
\caption{Passive: Gender and number breakdown}
\label{tab:passive}
\centering
\begin{tabular}{|l|l|l|}
\hline
\multicolumn{1}{|c|}{\textbf{Case name}} & \multicolumn{1}{c|}{\textbf{P@1}} & \multicolumn{1}{c|}{\textbf{P@10}} \\ \hline
Feminine Singular                               & 17.86\%                           & 50.71\%                            \\ \hline
Masculine Singular                               & 95.89\%                           & 89.32\%                            \\ \hline
Feminine Plural                               & 100.00\%                          & 55.00\%                            \\ \hline
Masculine Plural                               & 98.53\%                           & 81.47\%                            \\ \hline
\end{tabular}
\end{table}

\subsubsection{Discussion}

Summing up the results of the grammatical tests, we can see in Table \ref{tab:overall} that tasks that require no agreement had the best results, with 100\% precision for connectors and 98.78\% for impersonal verbs. Where morphsyntactic characteristics of the language played a role, we see that the results fall below 90\% even considering only the candidate in which the model has the most confidence. 

Interestingly, the nominal agreement test was the only one that shows better results among the top 10 candidates in comparison to the top 1. This could possibly mean that, for selecting adjectives, lexical cues on the context are stronger than the morphosyntactic information in the cue word. Considering the results reported by \cite{bacon2019does} for Portuguese on mBERT, we see that the model here had much worse performance for nominal agreement. This could be because we evaluated all candidates, and not only the ones that had the same part-of-speech as the masked word.

As the only closed class task, we expected the largest difference  from the evaluation among the top 1 to the top 10 candidates for connectors, as the model would eventually run out of options to fit in the context. However, the worse performance overall was seen in the task of verb agreement, where the model had problems finding good candidates for a few personal pronouns, both at the top 1 and at the top 10.

Moreover, frequency also seems to play a role in model performance, and prediction accuracy decreases for less frequent forms, including very formal pronouns or those that are frequently omitted. 

\begin{table}[h!]
\caption{Summary of Grammatical Proficiency Tests}
\label{tab:overall}
\centering
\begin{tabular}{|l|l|l|}
\hline
\multicolumn{1}{|c|}{\textbf{Test}} & \multicolumn{1}{c|}{\textbf{P@1}} & \multicolumn{1}{c|}{\textbf{P@10}} \\ \hline
Nominal Agreement                    & 77.53\%                           & 89.67\%                            \\ \hline
Verb Agreement                     & 79.56\%                           & 59.55\%                            \\ \hline
Subject Agreement                 & 83.38\%                           & 75.70\%                            \\ \hline
Connectors                              & 100.00\%                          & 54.17\%                            \\ \hline
Impersonal Verbs                       & 98.78\%                           & 86.77\%                            \\ \hline
Passive                             & 86.65\%                           & 78.38\%                            \\ \hline
\end{tabular}
\end{table}

\section{Conclusions and Future Work}\label{sec:conclusion}

In this work we addressed the problem of intrinsic evaluation of large scale language models and proposed a battery of model-agnostic tests to assess linguistic proficiency in Portuguese. This paper focused on some widely adopted neural language models of Brazilian Portuguese, and evaluated their performance in lexical, syntactic and semantic tasks. We developed a dataset with evaluation items that cover two particular areas: MWE inventory of idiomatic items, and proficiency in 6 grammatical tasks.  

Overall, the larger and language-specific models performed better in the MWE task. Although mBERT shows good capacities in different NLP tasks, the model captured little to no information to predict idiomatic items of Brazilian Portuguese. Despite the small difference in performance between BERTimbau Large and Base, the larger version presented a better recognition of MWE. However, exploring more outputs from the BERTimbau Base might have the same performance as BERTimbau Large.


The grammatical tests showed that BERTimbau Large has a good overall precision 
in the generation of candidates for masked items, especially when we looked only at the top 1 candidates, going up (or very close) to 100\% precision both for connectors and impersonal verbs. Even so, for tasks that required morphosyntactic agreement, there was a fall in precision, with the worse results (below 80\% among the top 1 candidates) being reported for nominal and verb agreement. The case of verb agreement was especially challenging for the model, because it consistently failed to produce good results for certain personal pronouns (first person plural, and second person singular and plural), which could be a sign of poor morphosyntactic generalisation, or be a side effect of the training corpus.

We adopted two evaluation criteria, one more strict, considering only the best candidate, and a more lenient one, which includes the top 10 candidates. Moreover, by considering not only the expected target word forms during the evaluation, but also considering alternative, grammatically correct outputs, we were able to detect the capacity of the model to produce different types of word forms for different contexts. Although deviant word forms did not represent much of the correct responses, amounting to around 5\% in a few tasks, they showed that a syntactic cue might not be as strong as the overall context of the sentence, as argued by \cite{gulordava2018colorless}. The results obtained confirm that the model achieves proficiency levels in tasks that do not require morphosyntactic agreement. However, it still lacks quality in certain items, in particular related to feminine singular forms. We also observed that there are instances (e.g. nominal agreement) in which the model has higher confidence (i.e. higher probability) in inadequate responses.
All these evaluations led to a profile of the model's linguistic information.  


As future work, we intend to extend the battery of tests to other linguistic aspects, such as selectional preferences for verbs, and inventory of collocations and terminology. We also intend to investigate whether possible biases in the distribution of the training data can affect the performance of these patterns. 
Finally, we plan to develop a multilingual version of the test, adapting it to closely related languages that share some of these linguistic patterns, and assessing if language proximity can be beneficial for few-shot learning scenarios.  


\begin{acknowledgements}
 This work has been developed in the framework of the project COURAGE (no. 95567), funded by the Volkswagen Foundation in the topic Artificial Intelligence and the Society of the Future. 
 It is also partly funded by the EPSRC project MIA: Modeling Id-iomaticity in Human and Artificial Language Pro-cessing (EP/T02450X/1) and by Research England, in the form of the Expanding Excellence in England (E3) programme. 
\end{acknowledgements}

%
\section*{Conflict of interest}
 The authors declare that they have no conflict of interest.

\bibliographystyle{nlelike}      

\bibliography{references}

%
%

\end{document}